\documentclass{article}

\PassOptionsToPackage{numbers, compress}{natbib}
\usepackage{wrapfig}
\usepackage{lipsum}

\usepackage[final]{nips_2017}

\usepackage[utf8]{inputenc} 
\usepackage[T1]{fontenc}    
\usepackage{hyperref}       
\usepackage{url}            
\usepackage{booktabs}       
\usepackage{amsfonts}       
\usepackage{nicefrac}       
\usepackage{microtype}      
\usepackage{graphicx}
\usepackage{amssymb}
\usepackage{amsmath}
\usepackage{multirow}

\title{Multi-Stage Variational Auto-Encoders for Coarse-to-Fine Image Generation}

\author{
  Lei Cai \\
  Washington State University\\
  Pullman, WA 99164 \\
  \texttt{lei.cai@wsu.edu} \\
  \And
  Hongyang Gao\\
  Washington State University\\
  Pullman, WA 99164 \\
  \texttt{hongyang.gao@wsu.edu} \\
  \And
  Shuiwang Ji \\
  Washington State University\\
  Pullman, WA 99164 \\
  \texttt{sji@eecs.wsu.edu} \\
}

\begin{document}

\maketitle

\begin{abstract}
Variational auto-encoder (VAE) is a powerful unsupervised
learning framework for image generation. One drawback of VAE is that
it generates blurry images due to its Gaussianity assumption and
thus $\ell_{2}$ loss. To allow the generation of high quality images
by VAE, we increase the capacity of decoder network by employing
residual blocks and skip connections, which also enable efficient
optimization. To overcome the limitation of $\ell_{2}$ loss, we
propose to generate images in a multi-stage manner from coarse to
fine. In the simplest case, the proposed multi-stage VAE divides the
decoder into two components in which the second component generates
refined images based on the course images generated by the first
component. Since the second component is independent of the VAE
model, it can employ other loss functions beyond the $\ell_{2}$ loss
and different model architectures. The proposed framework can be
easily generalized to contain more than two components. Experiment
results on the MNIST and CelebA datasets demonstrate that the
proposed multi-stage VAE can generate sharper images as compared to
those from the original VAE.
\end{abstract}

\section{Introduction}

In recent years, progress in deep learning has promoted the development of
generative models\cite{goodfellow2014generative,van2016wavenet,dinh2016density,hinton1986learning,bengio2014deep} that are able to capture the distributions
of high-dimensional dataset and generate new samples. Variational auto-encoder
(VAE)\cite{rezende2014stochastic} is a powerful unsupervised learning
framework for deep generative modeling. In VAE, the input data is encoded into
latent variables before they are reconstructed by the decoder network. The VAE
learns the transformation parameters by optimizing a variational lower bound
of the true likelihood. The lower bound consists of two components. The first
component is the Kullback-Leibler (KL) divergence between the approximate
posterior and a prior distribution, which is commonly a normal distribution.
The second component is the reconstruction loss given a latent variable. The
VAE assumes that the output follows a normal distribution given the latent
variable, thereby leading to an $\ell_{2}$ loss in the objective function. It
has been shown that the $\ell_{2}$ loss leads to blurry images when the data
are drawn from multi-modal distributions.

To make the VAE generate high quality images, some approaches have been
proposed to improve the decoder network
\cite{gulrajani2016pixelvae,kingma2016improving,chen2016variational}.
Since the decoder network is usually implemented with convolutional
neural networks (CNNs) \cite{lecun1998gradient}, we can increase the
network depth to improve the capacity of decoder networks as in
\cite{krizhevsky2012imagenet,simonyan2014very,szegedy2015going}.
However, deeper networks can be difficult to optimize. Therefore, we
employ the deep residual blocks, which are easy to optimize, to
increase the capacity of decoder. By employing residual blocks in
the decoder network, the VAE can generate high quality images.
However, it still suffers from the effect of $\ell_{2}$ loss and
thus generates blurry images.

\begin{figure}[t]
\centering
\includegraphics[width=\textwidth]{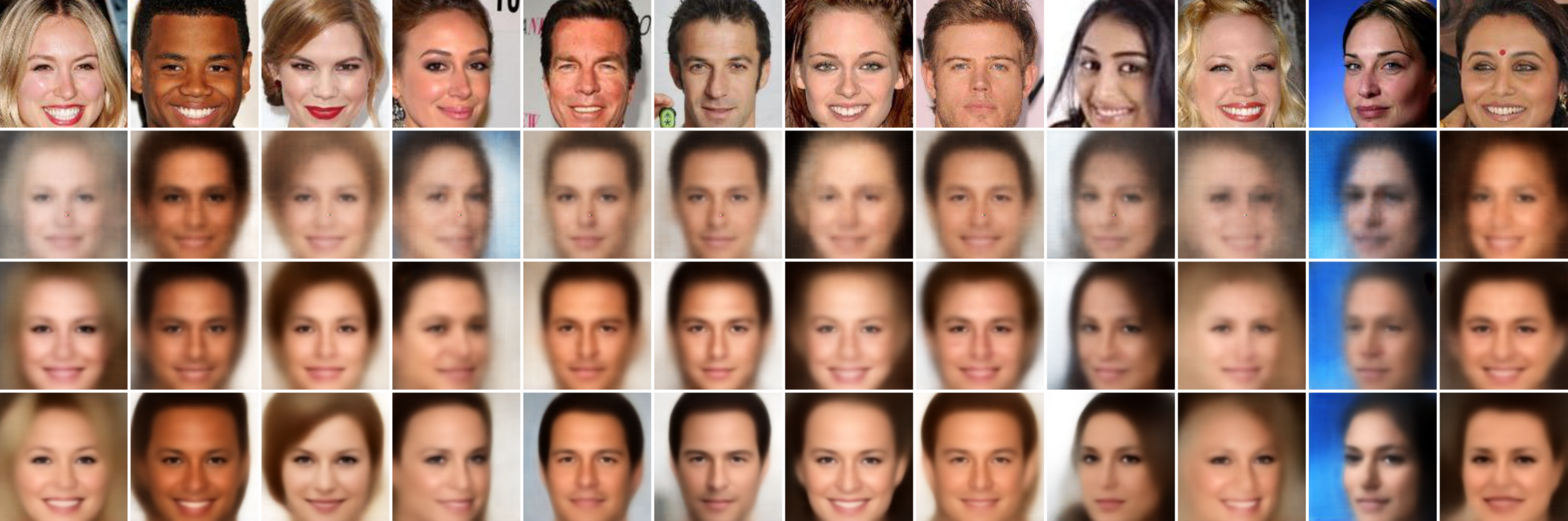}
\caption{Comparison of reconstructed images from the CelebA dataset.
The first row is the input images in the CelebA training set. The
second row is the reconstructed images generated by the original
VAE. The third and fourth rows are the results of deep residual VAE
and multi-stage VAE, respectively.} \label{fig:intro_res}
\end{figure}

In this work, we propose a multi-stage VAE framework to generate
high quality images. The key idea of multi-stage VAE is to generate
images from coarse to fine. One challenge is that, since the decoder
network is trained end-to-end, it is difficult to control the
decoder network and make it generate images from coarse to fine. A
simple solution is to train two models separately in which the first
model generates a coarse image and the second model refines the
coarse image. A drawback of this simple approach is that it reduces
the efficiency of the model and involves more computational costs.
To obtain fine images efficiently, we propose to employ an
$\ell_{2}$ loss in the middle of the decoder network, thus requiring
coarse images to be generated in an intermediate stage of the
decoder network. The remaining parts of the encoder network can be
considered as a model that takes coarse images as inputs and
generates refined versions of them as outputs. Indeed, the second
network can be considered as a super-resolution network. Following
this interpretation, we can employ any loss functions to refine the
images in the super-resolution network\cite{kim2016accurate}, thereby overcoming the
effect of $\ell_{2}$ loss. In this way, we can generate images from
coarse to fine and alleviate the effect of $\ell_{2}$ loss without
introducing extra parameters. Experimental results on the MNIST and
CelebA datasets demonstrate that the proposed multi-stage VAE can
capture more details and generate sharper images than the original
VAE. Some sample results are given in Figure~\ref{fig:intro_res}



\section{Multi-Stage Variational Auto-Encoder}

\subsection{Variational Auto-Encoder}

Variational auto-encoder (VAE) \cite{kingma2013auto} is a generative
model that is able to capture the probability distribution over
high-dimensional datasets. For image generation tasks, given a
dataset $X=\{x^{(i)}\}_{i=1}^{N}$, we wish to learn a distribution
function that can capture the dependencies among pixels. To tackle
this problem, we can train a distribution model $p_{\theta_{1}}(x)$,
parameterized by $\theta_{1}$, to approximate the data distribution
and optimize the model by maximizing the log likelihood as follows:
\begin{equation}
\log p_{\theta_{1}}(X) = \log p_{\theta_{1}}(x^{(1)},\ldots , x^{(N)}) = \sum_{i=1}^{N} \log p_{\theta_{1}}(x^{(i)}).
\end{equation}

However, probability distributions in high-dimensional space are
very difficult to model. Thus, a low-dimensional latent variable $z$
is usually introduced. It has been shown in \cite{kingma2013auto}
that the latent variable models can be optimized efficiently by
maximizing a variational lower bound on the likelihood function as
\begin{equation}
\log p_{\theta_{1}}(x) \geq \mathbb{E}_{q_{\phi}(z|x)}[\log p_{\theta_{1}}(x|z)]-D_{KL}[q_{\phi}(z|x)|p_{\theta_{1}}(z)]=-\mathcal{L}_{VAE},
\end{equation}
where $\mathcal{L}_{VAE}$ is the loss function we need to minimize
in VAE, and $q_{\phi}(z|x)$ is an approximate representation of the
intractable $p_{\theta_{1}}(z|x)$ parameterized by $q_{\phi}$. The
output distribution in the first term is often Gaussian as:
\begin{equation}
p_{\theta_{1}}(x|z) =  \mathcal{N} (x;f_{\theta_{1}}(z),\sigma^{2}I)
=C\times\exp\left(-\frac{(x-f_{\theta_{1}}(z))^{2}}{2\sigma
^{2}}\right),
\end{equation}
where $C$ is a constant, and $f_{\theta_{1}}(\cdot)$ is computed by
CNNs \cite{krizhevsky2012imagenet}. Therefore, the log likelihood
can be expressed as:
\begin{eqnarray}
\log p_{\theta_{1}}(X|z)
&= & \sum_{i=1}^{N} \log C\times\exp\left(-\frac{(x^{(i)}-f_{\theta_{1}}(z^{(i)}))^{2}}{2\sigma ^{2}}\right), \\
&= & N\times C- \frac{1}{2\sigma^{2}}
\sum_{i=1}^{N}(x^{(i)}-f_{\theta_{1}}(z^{(i)}))^2,
\end{eqnarray}
where $ N\times C$ is a constant that is irrelevant to
$f_{\theta_{1}}(\cdot)$ and can be ignored in optimization. The
first term in $\mathcal{L}_{VAE}$ is a $\ell_{2}$ loss between $x$
and $f_{\theta_{1}}(z)$. The second term corresponds to the
Kullback-Leibler (KL) divergence between $q_{\phi}(z|x)$ and
$p_{\theta_{1}}(z)$. VAE assumes that
$q_{\phi}(z|x)=\mathcal{N}(z;\mu_{\phi}(x),\sum_{\phi}(x))$ and
$p_{\theta}(z)=\mathcal{N}(z;0,I)$. $\mu_{\phi}(x)$ and
$\sum_{\phi}(x)$ are also implemented by CNNs. The second term in
$\mathcal{L}_{VAE}$ can be considered as a prior regularization.
Therefore, the loss function of VAE can be written as
\begin{equation}
\mathcal{L}_{VAE} = \mathcal{L}_{\ell_{2}} + \mathcal{L}_{prior},
\end{equation}
where
\begin{eqnarray}
\mathcal{L}_{\ell_{2}} &=& -\mathbb{E}_{q_{\phi}(z|x)}[\log p_{\theta_{1}}(x|z)]=  \frac{1}{2\sigma^{2}} \sum_{i=1}^{N}(x^{(i)}-f_{\theta_{1}}(z^{(i)}))^2,\\
\mathcal{L}_{prior} &=& D_{KL}[q_{\phi}(z|x)|p_{\theta_{1}}(z)].
\end{eqnarray}

\subsection{Deep Residual Variational Auto-Encoder}

VAE has shown promising results in image generation tasks\cite{dosovitskiy2016generating,zhao2017towards,larsen2016autoencoding}. However,
the images generated by VAE are blurry. This is caused by the
$\ell_{2}$ loss, which is based on the assumption that the data
follow a single Gaussian distribution. When samples in dataset have
multi-modal distribution, VAE cannot generate images with sharp
edges and fine details. In VAE, images are generated by
$f_{\theta_{1}}(\cdot)$. It is possible to generate better images by
using more complex model for $f_{\theta_{1}}(\cdot)$.  One solution
is to employ the autoregressive model \cite{oord2016pixel}\cite{van2016conditional} for
decoder function $f_{\theta_{1}}(\cdot)$. In the autoregressive
model, each pixel is conditioned on previously generated pixels. The
autoregressive model increases the dependency between pixels and
generates images with fine details. However, since it must generate
images pixel by pixel, the prediction procedure of autoregressive
model is much slower compared with other generative models such as
VAE.

\begin{figure}[t]
  \centering
  \includegraphics[width=\textwidth]{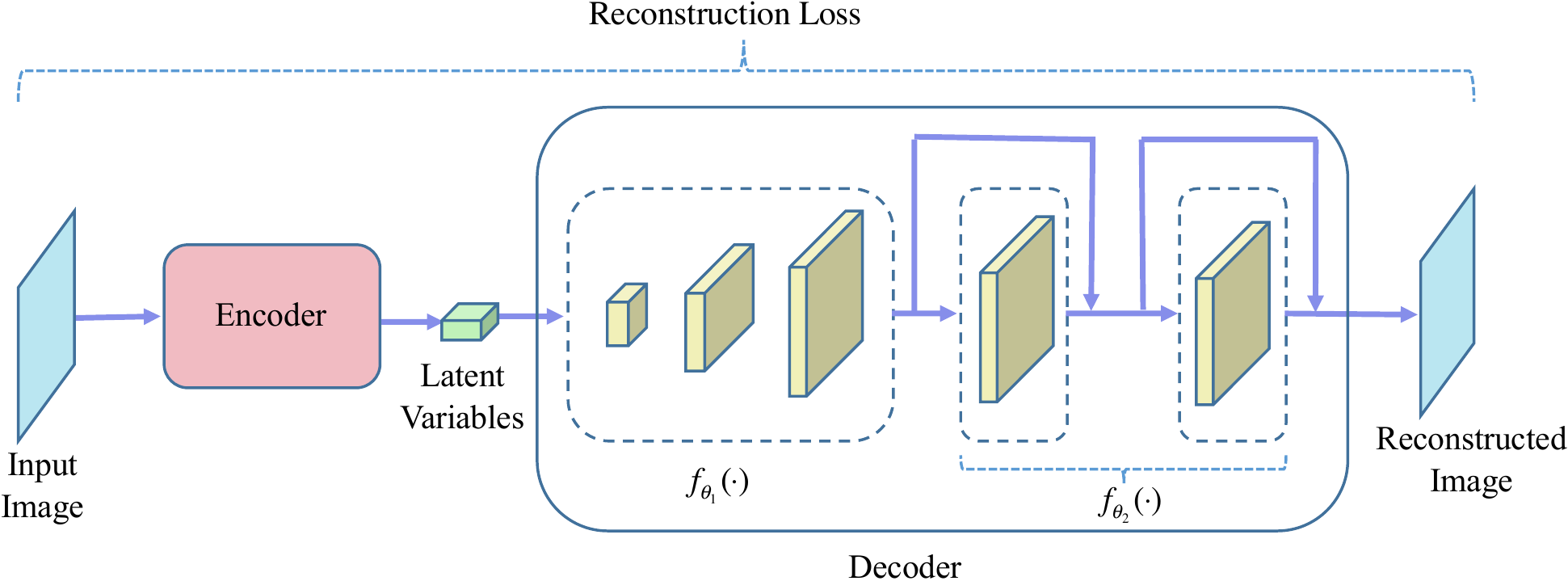}
  \caption{The network architecture of deep residual VAE. In this model, the encoder takes images as input and generate
  latent variables. The latent variables are fed into decoder network to recover
  the original spatial information. To make the decoder generate better image,
  we concatenate the original decoder network $f_{\theta_{1}}(\cdot)$ with the residual
  block network $f_{\theta_{2}}(\cdot)$ to increase the capacity of model.}
  \label{fig:RSVAE}
\end{figure}

Since the decoder of VAE is implemented with CNNs, a direct way to
generate better images is to employ deeper networks, resulting in
increased capacity of the decoder model \cite{simonyan2014very}. The
difficulty that deep neural networks facing is the degradation
problem. As the network depth increases, the performance of deep
networks initially improves and then degrades rapidly. Although deep
neural network models with higher capacity usually yield better
performance, it is also challenging to optimize them. To efficiently
train deep neural networks, the batch normalization method is
proposed in \citet{ioffe2015batch} by reducing internal covariate
shift. Another solution is the residual learning framework proposed
in \citet{he2016deep}, which employs the residual blocks and skip
connection to back-propagate the gradients more efficiently in the
network. The introduction of skip connection and residual block
makes the optimization of deep neural networks more efficient. It is
possible to employ deeper neural works on complex tasks. The
residual learning framework has already been successfully applied to
image recognition, object detection, and image super-resolution. To
increase the capacity of decoder in VAE and optimize the model
efficiently, we concatenate the original VAE decoder with several
residual blocks. The architecture of deep residual VAE is
illustrated in Figure \ref{fig:RSVAE}. Given the original decoder
$f_{\theta_{1}}(z)$, the deeper decoder networks can be denoted as
$f_{\theta}=f_{\theta_{2}}(f_{\theta_{1}}(z))$, where
$f_{\theta_{2}}(\cdot)$ corresponds to the residual network.
Compared with the original VAE decoder, the deeper decoder networks
can capture more details. The loss function of deep residual VAE can
be written as:
\begin{equation}
\mathcal{L}_{RSVAE} =\mathcal{L}_{\ell_{2}} + \mathcal{L}_{prior},
\end{equation}
where
\begin{eqnarray}
\mathcal{L}_{\ell_{2}} &=& -\mathbb{E}_{q_{\phi}(z|x)}[\log
p_{\theta_{1}}(x|z)]=  \frac{1}{2\sigma^{2}}
\sum_{i=1}^{N}(x^{(i)}-f_{\theta_{2}}(f_{\theta_{1}}(z^{(i)})))^2,\\
\mathcal{L}_{prior}& =& D_{KL}[q_{\phi}(z|x)|p_{\theta_{1}}(z)] .
\end{eqnarray}

\subsection{Multi-Stage Variational Auto-Encoder}

Experiment results in Section 3 show that deep residual VAE can
capture more details than the original VAE by adding residual blocks
to the decoder network. But the performance of deep residual VAE
saturates rapidly as more residual blocks are added. As the depth of
decoder network increases, the quality of generated images improves
with smaller and smaller margins. This saturation effect is not a
surprise as the network still employs $\ell_{2}$ loss and thus
generates blurry images. On the other hand, it is natural to use a
step-by-step procedure to generate high-quality images.
Specifically, in image generation, we can generate a coarse image
with rough shape and basic colors first and then refine the coarse
image to a high quality one. In VAE, the decoder network is trained
end-to-end. Thus we cannot control the process of image generation.
To make the decoder network generate images step-by-step, we need to
divide the decoder network into two components, where the first
component generates a coarse image, and the second component refines
it to a high quality one. To achieve this, we propose to add a loss
function at some location in the decoder network and enforce the
network to generate images at that location.

\begin{figure}[t]
\centering
\includegraphics[width=\textwidth]{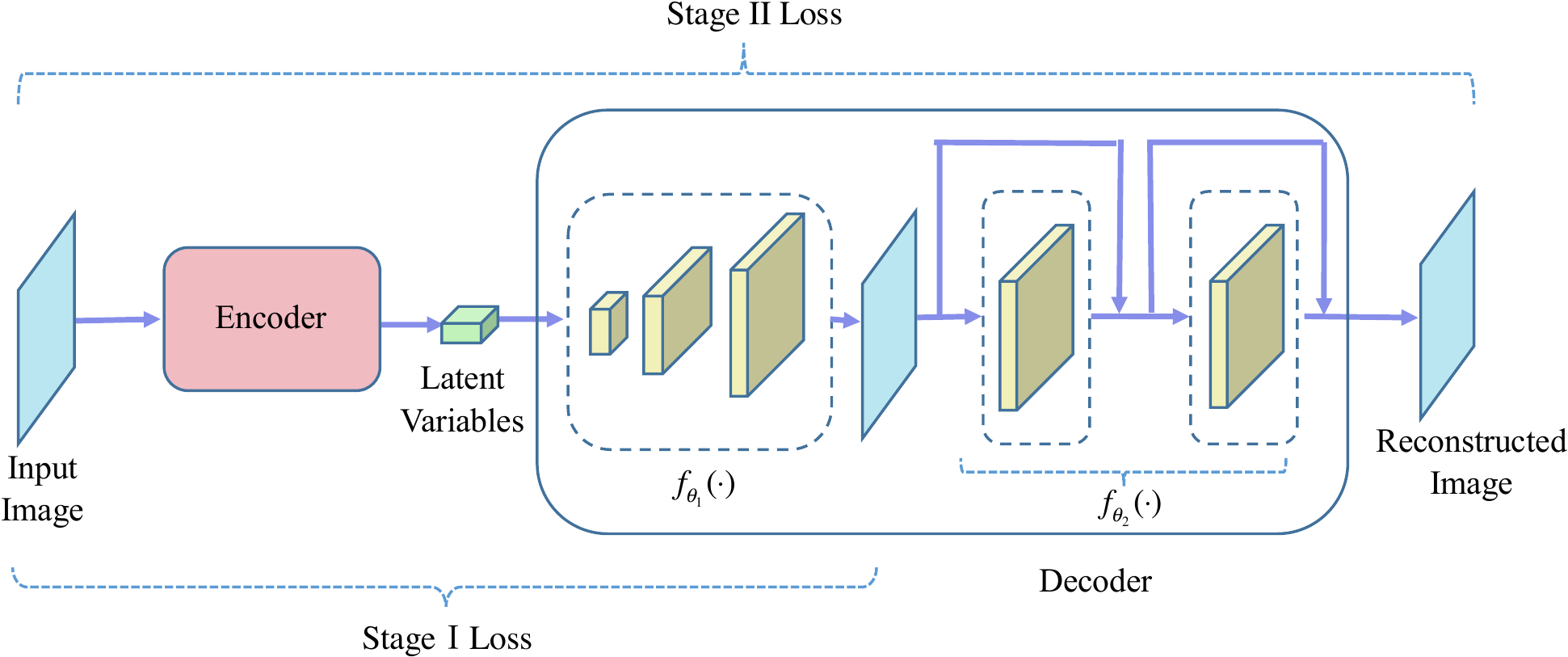}
\caption{The network architecture of multi-stage VAE based on the
deep residual VAE. In the first stage, the sub-network
$f_{\theta_{1}}(\cdot)$ generates a coarse image
$f_{\theta_{1}}(z)$. In the second stage, the coarse image
$f_{\theta_{1}}(z)$ is fed into the model $f_{\theta_{2}}(\cdot)$ to
produce a fine image $f_{\theta_{2}}(f_{\theta_{1}}(z))$.}
\label{fig:msvae}
\end{figure}

Here we use two stage deep VAE to illustrate how this idea works.
Since in the first stage we only need to generate a coarse image, it
is possible for the original VAE to accomplish this using the decode
function $f_{\theta_{1}}(\cdot)$. Then we need to build a model to
refine the coarse images. When we require the sub-network
$f_{\theta_{1}}(\cdot)$ in the decoder of deep residual VAE to
generate a coarse image, the input of $f_{\theta_{2}}(\cdot)$ is not
some arbitrary intermediate feature maps but a coarse image. In this
way, the sub-network $f_{\theta_{2}}(\cdot)$ acts as a model to
refine the coarse images generated from $f_{\theta_{1}}(z)$. The
architecture of the proposed multi-stage VAE is illustrated in
Figure \ref{fig:msvae}. The loss of the multi-stage VAE can be
written as:
\begin{equation}\label{eq: msvae}
\mathcal{L}_{MSVAE} = - \mathbb{E}_{q_{\phi}(z|x)}[\log
p_{\theta}(x|z)]+D_{KL}[q_{\phi}(z|x)|p_{\theta}(z)] +
\mathcal{L}_{rf}(x,f_{\theta_{2}}(f_{\theta_1}(z))).
\end{equation}

Compared with deep residual VAE, multi-stage VAE has two cost
functions in the decoder network. The cost function of the first
stage corresponds to $-\mathbb{E}_{q_{\phi}(z|x)}[\log
p_{\theta}(x|z)]$ in the original VAE, and it is used to generate
coarse images. The cost function of the second stage corresponds to
the third term in Equation~\ref{eq: msvae}, and it is used to refine
the coarse images. In multi-stage VAE framework, the second network
is independent of the VAE model. Therefore, we  can employ loss
function on $\mathcal{L}_{rf}(x,f_{\theta_{2}}(f_{\theta_1}(z)))$.
It also overcomes the effect of $\ell_{2}$ loss under the assumption
that data have a single Gaussian distribution. By employing
different loss functions, the second model can recover more detailed
information from blurry images. The $\mathcal{L}_{MSVAE}$ can be
written as:
\begin{equation}
\mathcal{L}_{MSVAE} = \mathcal{L}_{\ell_{2}} + \mathcal{L}_{prior} +
\mathcal{L}_{rf}(x,f_{\theta_{2}}(f_{\theta_1}(z))),
\end{equation}
where
\begin{eqnarray}
\mathcal{L}_{l_{2}} &=& -\mathbb{E}_{q_{\phi}(z|x)}[\log p_{\theta_{1}}(x|z)]=  \frac{1}{2\sigma^{2}} \sum_{i=1}^{N}(x^{(i)}-f_{\theta_{1}}(z^{(i)}))^2 ,\\
\mathcal{L}_{prior} &=& D_{KL}[q_{\phi}(z|x)|p_{\theta_{1}}(z)].
\end{eqnarray}

In addition, generating higher resolution images (e.g.,
$128\times128$) is challenging for generative models. In multi-stage
VAE, the coarse images generated in the first stage provide
additional information and subsequently enables the multi-stage VAE
to generate high-resolution images. The idea of tackling complex
tasks in a multi-stage manner is also employed by Stack GAN
\cite{zhang2016stackgan}. Stack GAN employs two separate models to
generate low-resolution images and high-resolution images,
respectively. The two models are trained separately. However, our
model divides the decoder network into two components with different
loss functions, and both networks are trained jointly.

\subsection{Connections with Super-Resolution}

We employ residual networks in the second stage of our multi-stage
VAE to refine the coarse images generated in the first stage. The
key idea of the second model is similar to the super-resolution
residual net (SRResNet) \cite{ledig2016photo}. In SRResNet, a
low-resolution image is fed into a network composed of residual
blocks and up-sampling layers. Then an image with high resolution is
generated by SRResNet.

In multi-stage VAE, we employ a pixel-wise loss function to recover
the details between low-resolution images and high-resolution
images. Minimizing the pixel-wise loss encourages the model to
generate the average of plausible solutions, thus leading to poor
perceptual quality \cite{dosovitskiy2016generating,mathieu2015deep}.
A plausible loss function applied in image super-resolution tasks is
the combination of Euclidean distances in feature space and
adversarial loss. In fact, our multi-stage VAE framework can work
with any plausible super-resolution model by replacing the loss
function in $\mathcal{L}_{rf}$ and the model architecture of
$f_{\theta_{2}}(\cdot)$.

\section{Experiments}

In this section, we evaluate the deep residual VAE and multi-stage
VAE on the MNIST and CelebA datasets and compare the quality of
generated images with the original VAE. Results show that the
proposed multi-stage VAE generates higher-resolution images as
compared to those generated by the original VAE and deep residual
VAE.

\subsection{Settings}

CelebA \cite{liu2015faceattributes} is a large scale face dataset that
contains $202,599$ face images. The size of each face image is 178$\times$218.
Most prior VAE work using this dataset crops the images to $64\times 64$. In
order to demonstrate the performance of our multi-stage VAE in generating
high-resolution images, we crop the image to 128$\times$128. We train three
models for 200000 iterations. with batch size of 32 and a learning rate of
2e-4. The encoder model of VAE consists of four layers. Each layer consists of
a convolution layer with stride 1 followed by a convolution layer with stride
2. The latent variable size of VAE is 512 for the CelebA dataset. The decoder
network consists of four deconvolution layers. To generate images with high
quality, five residual blocks are employed in the decoder network. The
$\ell_{1}$ loss is used in the objective function of the second network.

The MNIST is a handwritten digits dataset where the size of each
image is 28$\times$28. We train three models on the training set of
$60,000$ images. Each model is trained for $100,000$ iterations with
a batch size of 256 and a learning rate of 1e-3. The encoder model
of VAE consists of three convolution layers with a stride of $2$. The latent variable size of VAE is
set to $128$. The decoder reconstructs the image from the latent
variable with three deconvolution layers. To increase the complexity
of decoder network, we concatenate the original VAE decoder with
five residual blocks in the deep residual network. Each residual
block consists of two convolution layers followed by a batch
normalization layer. In multi-stage VAE, we add an $\ell_{2}$ loss
function at the location of the output in the original VAE. The
residual network is employed to refine the coarse images generated
in the first stage. To overcome the blurry effect of $\ell_{2}$
loss, we employ $\ell_{1}$ loss in the objective function of the
second network.

\subsection{Results and Analysis}

\begin{figure}[t]
\centering
\includegraphics[width=\textwidth]{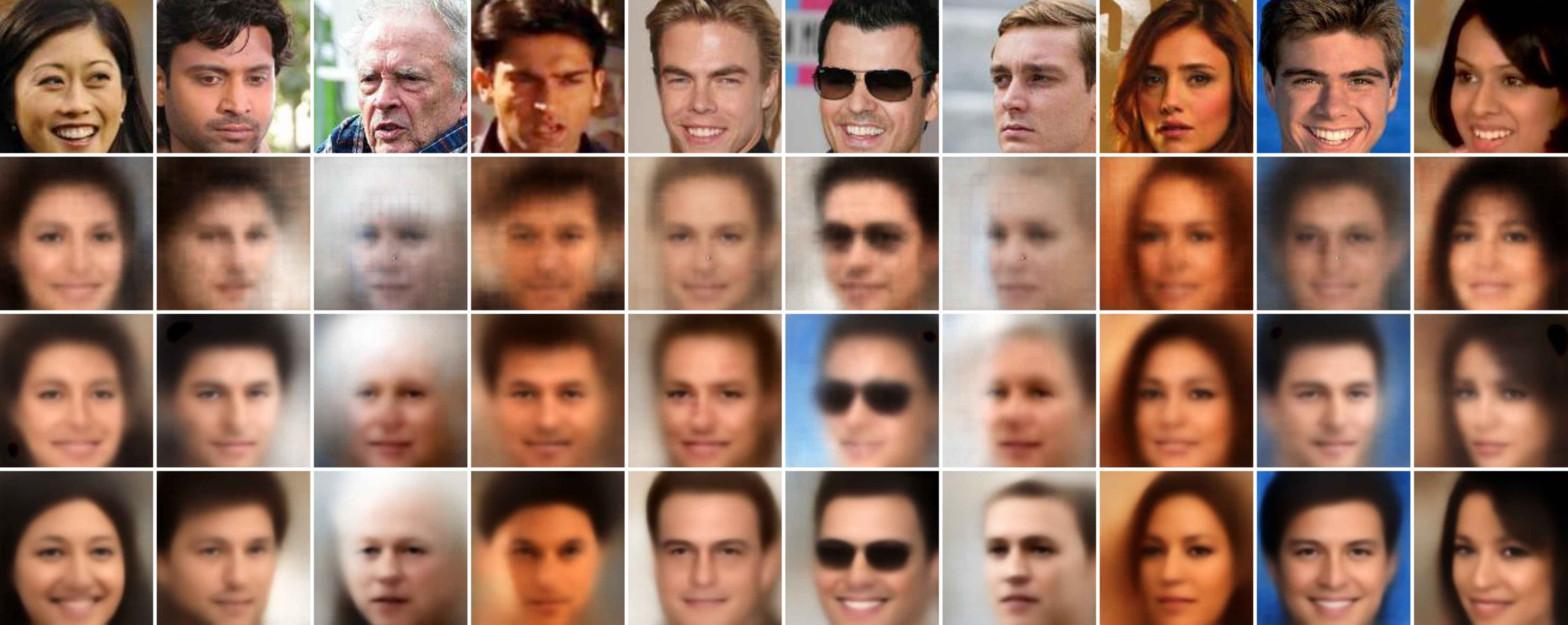}
\caption{Comparison of reconstructed images from the CelebA dataset.
The first row is the input images in the CelebA training set. The
second row is the reconstructed images generated by the original
VAE. The third and fourth rows are the results of deep residual VAE
and multi-stage VAE respectively.} \label{fig:celeba_res}
\end{figure}

\begin{figure}[t]
\centering
\includegraphics[width=\textwidth]{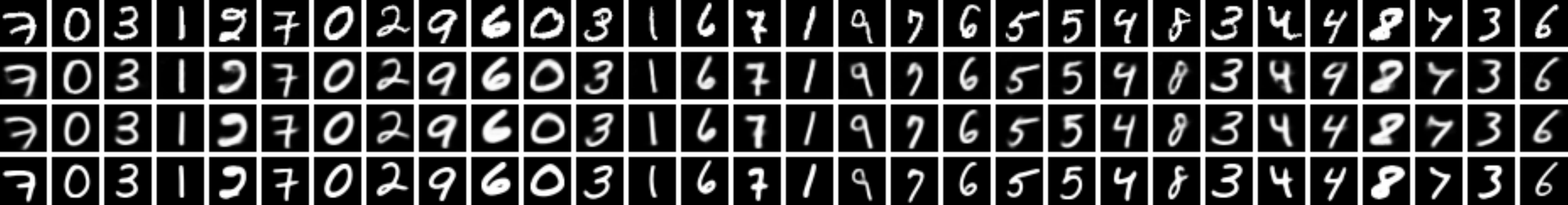}
\caption{Comparison of reconstructed images from the MNIST dataset.
The first row is the input images from the MNIST training set. The
second row is the reconstructed images generated by the original
VAE. The third and fourth rows are the results of deep residual VAE
and multi-stage VAE, respectively.} \label{fig:mnist_res}
\end{figure}

Figures \ref{fig:celeba_res} and \ref{fig:mnist_res} provide some
reconstructed images by different models. We can see that the deep
residual VAE can capture more details than the original VAE by
employing more complex decoder network. However, the images
generated by deep residual VAE are still blurry due to the effect of
$\ell_{2}$ loss. We also observe that the effect of $\ell_{2}$ loss
is largely overcome by employing the multi-stage loss. The blurry
region becomes clearer through the multi-stage refine process. These
results demonstrate that the proposed multi-stage VAE goes beyond
the bottleneck of increasing the capacity of decoder network,
thereby effectively overcoming the blurry effect caused by the
$\ell_{2}$ loss.

\begin{figure}[t]
\centering
\includegraphics[width=\textwidth]{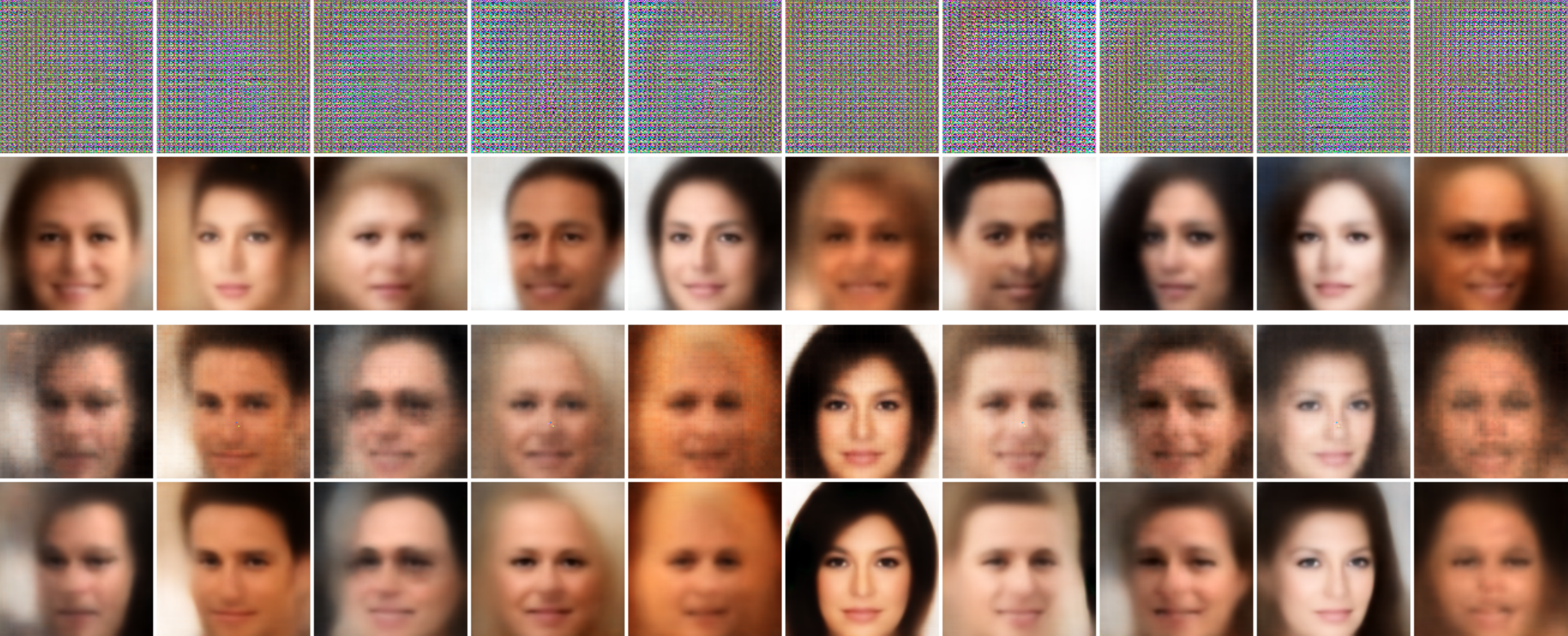}
\caption{Illustration of decoder outputs on the CelebA dataset. The
first and third rows are the output of $f_{\theta_{1}}(\cdot)$ in
deep residual VAE and multi-stage VAE, respectively. The second and
fourth rows are the outputs of $f_{\theta_{2}}(\cdot)$ in deep
residual VAE and multi-stage VAE, respectively.}
\label{fig:l_h_celeba}
\end{figure}
\begin{figure}[t]
\centering
\includegraphics[width=\textwidth]{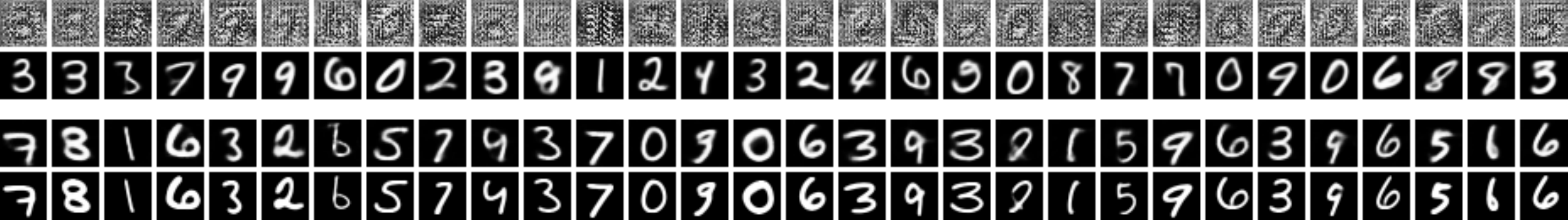}
\caption{Illustration of decoder output on the MNIST dataset. The
first and third rows are the outputs of $f_{\theta_{1}}(\cdot)$ in
deep residual VAE and multi-stage VAE, respectively. The second and
fourth rows are the outputs of $f_{\theta_{2}}(\cdot)$ in deep
residual VAE and multi-stage VAE, respectively.}
\label{fig:l_h_mnist}
\end{figure}

Figures \ref{fig:l_h_celeba} and \ref{fig:l_h_mnist} provide some
reconstructed images and intermediate outputs of
$f_{\theta_{1}}(\cdot)$ by the deep residual VAE and multi-stage
VAE. We can see that at the intermediate location in the decoder
network of multi-stage VAE, a blurry image is generated, and it is
fed into the residual networks. Through the refined operation of the
second network, an image with high quality is generated. Since the
whole decoder network of deep residual VAE only contains a single
loss function, the generation process suffers from the effect of
$\ell_{2}$ loss. Therefore, the images generated by deep residual
VAE are still blurry.

\begin{figure}[t]
\centering
\includegraphics[width=\textwidth]{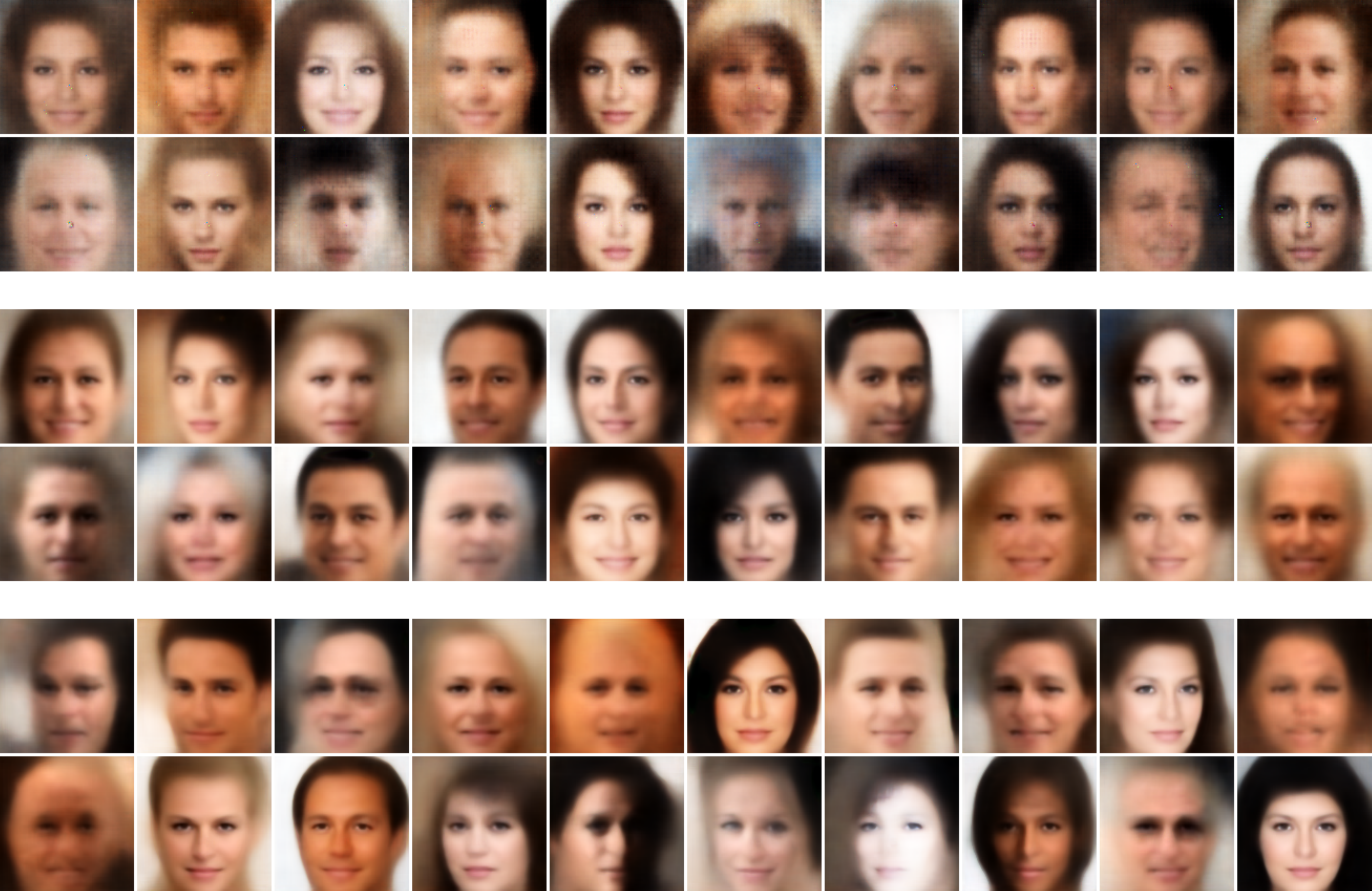}
\caption{Sample images generated by different models when trained on
the CelebA dataset. The first two rows are images generated by a
standard VAE. The middle two rows are images generated by deep
residual VAE. The last two rows are images generated by multi-stage
VAE.}\label{fig:celeba:pred}
\end{figure}

\begin{figure}[t]
\centering
\includegraphics[width=\textwidth]{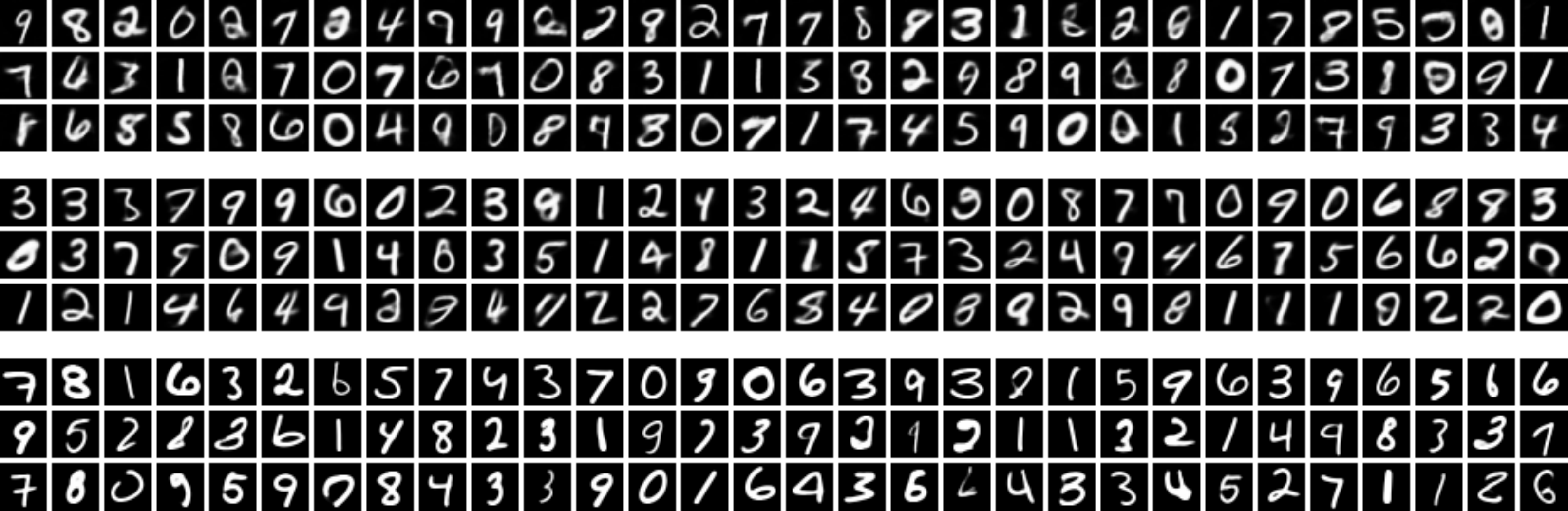}
\caption{Sample images generated by different models when trained on
the MNIST dataset. The first three rows are images generated by a
standard VAE. The middle three rows are images generated by deep
residual VAE. The last three rows are images generated by
multi-stage VAE.}\label{fig:mnist:pred}
\end{figure}

Figures \ref{fig:celeba:pred} and \ref{fig:mnist:pred} provide some
sample images generated by the original VAE, deep residual VAE, and
multi-stage VAE when the models are trained on the CelebA and MNIST
datasets. We can see that the images generated by the multi-stage
VAE have higher resolution than those generated by other two
methods. Also the images generated by the deep residual VAE are
clearer than those generated by the original VAE. These results
demonstrates that the proposed multi-stage VAE is effective in
generating high resolution images.

\section{Conclusion and Future Work}
In this work, we propose a multi-stage VAE that can generate higher
quality images than the original VAE. The original VAE always
generated blurry images due to the effect of $\ell_{2}$ loss. To
generate high quality images, we propose to improve the decoder
capacity by increasing the network depth and employing residual
blocks and skip connection. Although the deep residual VAE can
capture more details, it still suffers from the effect of $\ell_{2}$
loss and generates blurry images. To overcome the limitation of
$\ell_{2}$ loss, we propose to generate images from coarse to fine.
To achieve this goal, we require the decoder network to generate a
coarse image by employing a $\ell_{2}$ loss function in the first
stage. The subsequent stage in the decoder network acts as a
super-resolution network that takes a blurry image as input and
generates a high quality image. Since the super-resolution network
is independent of the VAE model, it can employ other loss functions
to overcome the the effect of $\ell_{2}$ loss, thereby generating
high quality images. Experimental results on the MNIST and CelebA
datasets show that the proposed multi-stage VAE can overcome the
effect of $\ell_{2}$ loss and generate high quality images.

One interpretation of our proposed framework is that, the network in
the second stage can be considered as a super-resolution module.
Following this interpretation, we plan to use other model
architectures and loss functions commonly used for super-resolution,
such as the adversarial loss~\cite{ledig2016photo}. As has been
mentioned, the proposed multi-stage framework can be generalized to
more than two components. We plan to explore more stages in the
future.

\subsubsection*{Acknowledgments}

This work was supported in part by National Science Foundation grants
IIS-1615035 and DBI-1641223, and by Washington State University. We gratefully
acknowledge the support of NVIDIA Corporation with the donation of the Tesla
K40 GPU used for this research.



%

%


{\scriptsize \bibliography{deep}}

\begin{thebibliography}{26}
\providecommand{\natexlab}[1]{#1}
\providecommand{\url}[1]{\texttt{#1}}
\expandafter\ifx\csname urlstyle\endcsname\relax
  \providecommand{\doi}[1]{doi: #1}\else
  \providecommand{\doi}{doi: \begingroup \urlstyle{rm}\Url}\fi

\bibitem[Bengio et~al.(2014)Bengio, Laufer, Alain, and
  Yosinski]{bengio2014deep}
Yoshua Bengio, Eric Laufer, Guillaume Alain, and Jason Yosinski.
\newblock Deep generative stochastic networks trainable by backprop.
\newblock In \emph{International Conference on Machine Learning}, pages
  226--234, 2014.

\bibitem[Chen et~al.(2016)Chen, Kingma, Salimans, Duan, Dhariwal, Schulman,
  Sutskever, and Abbeel]{chen2016variational}
Xi~Chen, Diederik~P Kingma, Tim Salimans, Yan Duan, Prafulla Dhariwal, John
  Schulman, Ilya Sutskever, and Pieter Abbeel.
\newblock Variational lossy autoencoder.
\newblock \emph{arXiv preprint arXiv:1611.02731}, 2016.

\bibitem[Dinh et~al.(2016)Dinh, Sohl-Dickstein, and Bengio]{dinh2016density}
Laurent Dinh, Jascha Sohl-Dickstein, and Samy Bengio.
\newblock Density estimation using real nvp.
\newblock \emph{arXiv preprint arXiv:1605.08803}, 2016.

\bibitem[Dosovitskiy and Brox(2016)]{dosovitskiy2016generating}
Alexey Dosovitskiy and Thomas Brox.
\newblock Generating images with perceptual similarity metrics based on deep
  networks.
\newblock In \emph{Advances in Neural Information Processing Systems}, pages
  658--666, 2016.

\bibitem[Goodfellow et~al.(2014)Goodfellow, Pouget-Abadie, Mirza, Xu,
  Warde-Farley, Ozair, Courville, and Bengio]{goodfellow2014generative}
Ian Goodfellow, Jean Pouget-Abadie, Mehdi Mirza, Bing Xu, David Warde-Farley,
  Sherjil Ozair, Aaron Courville, and Yoshua Bengio.
\newblock Generative adversarial nets.
\newblock In \emph{Advances in neural information processing systems}, pages
  2672--2680, 2014.

\bibitem[Gulrajani et~al.(2016)Gulrajani, Kumar, Ahmed, Taiga, Visin, Vazquez,
  and Courville]{gulrajani2016pixelvae}
Ishaan Gulrajani, Kundan Kumar, Faruk Ahmed, Adrien~Ali Taiga, Francesco Visin,
  David Vazquez, and Aaron Courville.
\newblock Pixelvae: A latent variable model for natural images.
\newblock \emph{arXiv preprint arXiv:1611.05013}, 2016.

\bibitem[He et~al.(2016)He, Zhang, Ren, and Sun]{he2016deep}
Kaiming He, Xiangyu Zhang, Shaoqing Ren, and Jian Sun.
\newblock Deep residual learning for image recognition.
\newblock In \emph{Proceedings of the IEEE Conference on Computer Vision and
  Pattern Recognition}, pages 770--778, 2016.

\bibitem[Hinton and Sejnowski(1986)]{hinton1986learning}
Geoffrey~E Hinton and Terrence~J Sejnowski.
\newblock Learning and releaming in boltzmann machines.
\newblock \emph{Parallel Distrilmted Processing}, 1, 1986.

\bibitem[Ioffe and Szegedy(2015)]{ioffe2015batch}
Sergey Ioffe and Christian Szegedy.
\newblock Batch normalization: Accelerating deep network training by reducing
  internal covariate shift.
\newblock \emph{arXiv preprint arXiv:1502.03167}, 2015.

\bibitem[Kim et~al.(2016)Kim, Kwon~Lee, and Mu~Lee]{kim2016accurate}
Jiwon Kim, Jung Kwon~Lee, and Kyoung Mu~Lee.
\newblock Accurate image super-resolution using very deep convolutional
  networks.
\newblock In \emph{Proceedings of the IEEE Conference on Computer Vision and
  Pattern Recognition}, pages 1646--1654, 2016.

\bibitem[Kingma and Welling(2013)]{kingma2013auto}
Diederik~P Kingma and Max Welling.
\newblock Auto-encoding variational bayes.
\newblock \emph{arXiv preprint arXiv:1312.6114}, 2013.

\bibitem[Kingma et~al.(2016)Kingma, Salimans, Jozefowicz, Chen, Sutskever, and
  Welling]{kingma2016improving}
Diederik~P Kingma, Tim Salimans, Rafal Jozefowicz, Xi~Chen, Ilya Sutskever, and
  Max Welling.
\newblock Improving variational autoencoders with inverse autoregressive flow.
\newblock In \emph{Advances In Neural Information Processing Systems}, pages
  4736--4744, 2016.

\bibitem[Krizhevsky et~al.(2012)Krizhevsky, Sutskever, and
  Hinton]{krizhevsky2012imagenet}
Alex Krizhevsky, Ilya Sutskever, and Geoffrey~E Hinton.
\newblock Imagenet classification with deep convolutional neural networks.
\newblock In \emph{Advances in neural information processing systems}, pages
  1097--1105, 2012.

\bibitem[Larsen et~al.(2016)Larsen, S{\o}nderby, Larochelle, and
  Winther]{larsen2016autoencoding}
Anders Boesen~Lindbo Larsen, S{\o}ren~Kaae S{\o}nderby, Hugo Larochelle, and
  Ole Winther.
\newblock Autoencoding beyond pixels using a learned similarity metric.
\newblock In \emph{Proceedings of The 33rd International Conference on Machine
  Learning}, pages 1558--1566, 2016.

\bibitem[LeCun et~al.(1998)LeCun, Bottou, Bengio, and
  Haffner]{lecun1998gradient}
Yann LeCun, L{\'e}on Bottou, Yoshua Bengio, and Patrick Haffner.
\newblock Gradient-based learning applied to document recognition.
\newblock \emph{Proceedings of the IEEE}, 86\penalty0 (11):\penalty0
  2278--2324, 1998.

\bibitem[Ledig et~al.(2016)Ledig, Theis, Husz{\'a}r, Caballero, Cunningham,
  Acosta, Aitken, Tejani, Totz, Wang, et~al.]{ledig2016photo}
Christian Ledig, Lucas Theis, Ferenc Husz{\'a}r, Jose Caballero, Andrew
  Cunningham, Alejandro Acosta, Andrew Aitken, Alykhan Tejani, Johannes Totz,
  Zehan Wang, et~al.
\newblock Photo-realistic single image super-resolution using a generative
  adversarial network.
\newblock \emph{arXiv preprint arXiv:1609.04802}, 2016.

\bibitem[Liu et~al.(2015)Liu, Luo, Wang, and Tang]{liu2015faceattributes}
Ziwei Liu, Ping Luo, Xiaogang Wang, and Xiaoou Tang.
\newblock Deep learning face attributes in the wild.
\newblock In \emph{Proceedings of International Conference on Computer Vision
  (ICCV)}, 2015.

\bibitem[Mathieu et~al.(2015)Mathieu, Couprie, and LeCun]{mathieu2015deep}
Michael Mathieu, Camille Couprie, and Yann LeCun.
\newblock Deep multi-scale video prediction beyond mean square error.
\newblock \emph{arXiv preprint arXiv:1511.05440}, 2015.

\bibitem[Oord et~al.(2016)Oord, Kalchbrenner, and Kavukcuoglu]{oord2016pixel}
Aaron van~den Oord, Nal Kalchbrenner, and Koray Kavukcuoglu.
\newblock Pixel recurrent neural networks.
\newblock \emph{arXiv preprint arXiv:1601.06759}, 2016.

\bibitem[Rezende et~al.(2014)Rezende, Mohamed, and
  Wierstra]{rezende2014stochastic}
Danilo~Jimenez Rezende, Shakir Mohamed, and Daan Wierstra.
\newblock Stochastic backpropagation and approximate inference in deep
  generative models.
\newblock In \emph{Proceedings of The 31st International Conference on Machine
  Learning}, pages 1278--1286, 2014.

\bibitem[Simonyan and Zisserman(2014)]{simonyan2014very}
Karen Simonyan and Andrew Zisserman.
\newblock Very deep convolutional networks for large-scale image recognition.
\newblock \emph{arXiv preprint arXiv:1409.1556}, 2014.

\bibitem[Szegedy et~al.(2015)Szegedy, Liu, Jia, Sermanet, Reed, Anguelov,
  Erhan, Vanhoucke, and Rabinovich]{szegedy2015going}
Christian Szegedy, Wei Liu, Yangqing Jia, Pierre Sermanet, Scott Reed, Dragomir
  Anguelov, Dumitru Erhan, Vincent Vanhoucke, and Andrew Rabinovich.
\newblock Going deeper with convolutions.
\newblock In \emph{Proceedings of the IEEE Conference on Computer Vision and
  Pattern Recognition}, pages 1--9, 2015.

\bibitem[van~den Oord et~al.(2016{\natexlab{a}})van~den Oord, Dieleman, Zen,
  Simonyan, Vinyals, Graves, Kalchbrenner, Senior, and
  Kavukcuoglu]{van2016wavenet}
A{\"a}ron van~den Oord, Sander Dieleman, Heiga Zen, Karen Simonyan, Oriol
  Vinyals, Alex Graves, Nal Kalchbrenner, Andrew Senior, and Koray Kavukcuoglu.
\newblock Wavenet: A generative model for raw audio.
\newblock 2016{\natexlab{a}}.

\bibitem[van~den Oord et~al.(2016{\natexlab{b}})van~den Oord, Kalchbrenner,
  Espeholt, Vinyals, Graves, et~al.]{van2016conditional}
Aaron van~den Oord, Nal Kalchbrenner, Lasse Espeholt, Oriol Vinyals, Alex
  Graves, et~al.
\newblock Conditional image generation with pixelcnn decoders.
\newblock In \emph{Advances in Neural Information Processing Systems}, pages
  4790--4798, 2016{\natexlab{b}}.

\bibitem[Zhang et~al.(2016)Zhang, Xu, Li, Zhang, Huang, Wang, and
  Metaxas]{zhang2016stackgan}
Han Zhang, Tao Xu, Hongsheng Li, Shaoting Zhang, Xiaolei Huang, Xiaogang Wang,
  and Dimitris Metaxas.
\newblock Stackgan: Text to photo-realistic image synthesis with stacked
  generative adversarial networks.
\newblock \emph{arXiv preprint arXiv:1612.03242}, 2016.

\bibitem[Zhao et~al.(2017)Zhao, Song, and Ermon]{zhao2017towards}
Shengjia Zhao, Jiaming Song, and Stefano Ermon.
\newblock Towards deeper understanding of variational autoencoding models.
\newblock \emph{arXiv preprint arXiv:1702.08658}, 2017.

\end{thebibliography}

\end{document}